# Building a European Multilingual Evaluation Dataset: The MMLU Localisation Project within the EMT Network

**Pilar Sánchez-Gijón**
Universitat Autònoma de Barcelona / Spain
pilar.sanchez.gijon@uab.cat

**Susana Valdez**
Leiden University/ the Netherlands
s.valdez@hum.leidenuniv.nl

**Sofía Calvo Del Barrio**
Universitat Autònoma de Barcelona / Spain
sofia.calvo@autonoma.cat

**Florence Bellemont**
Leiden University / The Netherlands
f.m.f.e.l.g.bellemont@umail.leidenuniv.nl

**Anna Kokkinidou**
Directorate-General for Translation of the European Commission / Belgium
anna.kokkinidou@ec.europa.eu

**Mihai Cristian Brasoveanu**
Directorate-General for Translation of the European Commission / Belgium
mihai.brasoveanu@ec.europa.eu

## Abstract

This paper reports on a collaboration between the Directorate-General for Translation (DGT) and the European Master's in Translation (EMT) to localise the MMLU dataset into 11 European languages. Beyond creating a more inclusive benchmark for LLM evaluation, the project offers master's students authentic, project-based professional training in translation, revision, project management, and multilingual coordination, while highlighting key methodological, administrative, and workflow challenges.

## 1 Introduction

The growing integration of artificial intelligence (AI) into language services has intensified demand for high-quality multilingual resources capable of evaluating AI systems across diverse linguistic contexts. One prominent example is the Massive Multitask Language Understanding (MMLU) benchmark (Hendrycks et al., 2021), a widely used dataset for assessing the reasoning and knowledge capabilities of large language models across 57 domains. While the MMLU dataset has become one of the leading benchmarks for evaluating Large Language Models (LLMs), several efforts have sought to create translated and culturally relevant versions into multiple languages (e.g., Global-MMLU (Singh et al. 2024) or MMMLU from OpenAI). These initiatives reflect a growing recognition that English-centric evaluation may pose significant limitations; yet most existing multilingual versions rely primarily on machine translation (MT) or crowdsourced post-editing, raising persistent questions about linguistic quality, terminological consistency, and cultural appropriateness in the target languages.

This paper reports on a large-scale localisation initiative that pursues a dual aim: to produce a human-translated version of the MMLU dataset into European languages, and to do so through an authentic, project-based training environment in which translators in training engage with an authentic institutional workflow. The initiative is carried out in collaboration between the network of European Master's in Translation (EMT) programmes and the Directorate-General for Translation (DGT) of the European Commission, and is coordinated within the EMT framework by the Working Group dedicated to Linguistic Data and Terminology. Beyond its technical objectives, the initiative serves as a structured, practice-based training environment in which master's students from participating universities engage in authentic multilingual workflows. In doing so, they contribute to the creation of a high-impact European language technology resource while simultaneously developing translation competences aligned with real-world translation scenarios.

This paper presents the design and pedagogical rationale of this ongoing initiative, and reflects on what this model of academia–institution collaboration may offer for translation training in the context of an increasingly AI-mediated language industry.

This paper is structured as follows: Section 2 reviews relevant literature on project-based learning. Section 3 reports on the DGT-EMT initiative, followed by a discussion of key challenges in section 4.

## 2 Didactical and Theoretical Background

In contemporary translation training, pedagogical approaches are increasingly grounded in holistic, learner-centred principles that closely mirror professional practice. Models such as Authentic Experiential Learning and Project-Based Learning (PjBL) emphasise real assignments and "learning by doing" (Korda, 2023, Kiraly 2005), while Situated Learning recreates professional environments—such as Simulated Translation Bureaus (STBs) (Van Egdom *et al.*, 2020)—in a structured setting. In this sense, these approaches operationalise the EMT Competence Framework by embedding professional competences into authentic tasks, while also drawing on social constructivist principles that position students as active participants engaged in collaborative problem-solving under guided facilitation.

The practical application of these pedagogical approaches can be seen in a number of successful projects. For instance, the INSTB Network (International Network of Simulated Translation Bureaus), established in 2015, is a strategic partnership that includes 19 institutions across 10 countries (Van Egdom *et al.*, 2020).

Project management within these training simulations mirrors professional practices and involves several critical steps. Students often take on key roles, such as project manager, translator, terminologist, QA manager, and reviser, all of which are vital to the functioning of the simulation. The creation of business plans, market assessments, and pricing strategies is also part of the learning process. Professional tools, such as Slack, are used to manage internal workflows, while students also engage in activities like "storytelling" for branding and creating websites to attract potential clients.

In authentic projects, students negotiate deadlines and clarify project briefs with external clients. In simulated projects, the instructor may take on the role of a "mock client," providing students with a controlled yet practical environment for learning client interaction.

The incorporation of real-world elements into these projects offers numerous benefits. Notably, these projects enhance employability by providing students with market-oriented experience, thus improving their ability to navigate the job market upon graduation (Uribe de Kellet, 2022). Additionally, the self-efficacy of students is significantly strengthened, as they develop confidence in managing professional risks and tasks. Moreover, soft skills such as teamwork, leadership, interpersonal communication, and negotiation are fostered through collaborative project work. Lastly, the international nature of these projects promotes internationalization, allowing students to engage in cross-cultural networking and develop intercultural competence.

However, these projects also present several challenges. External pressure from working on authentic projects can introduce unpredictability and the stress of real-world consequences. Aligning the logistics of academic calendars across different international partner universities can also prove to be a significant obstacle, hindering smooth collaboration. Additionally, ethical concerns arise in the context of STBs, as they must ensure that they do not undercut the market rates of professional translators. These simulations often focus on non-commercial or non-profit projects to avoid this issue. Furthermore, initial anxiety is common among students at the start of a project, given the demanding nature of professional standards. Finally, complexity is heightened when engaging in international collaborations, which may discourage some students, particularly when deadlines are tight.

## 3 The MMLU Dataset

### 3.1. Translation Project: The MMLU Dataset, setting the ground for the DGT-EMT project

Emerging from the need for a European benchmark for evaluating LLMs, this project was initiated by the AI Unit of the DGT in November 2025. It was developed within the framework of an ongoing collaboration between the DGT and the EMT network, bringing together institutional expertise in multilingual communication and professional translator training across participating European universities. In November 2025, the AI Unit of the DGT proposed a new collaboration to Working Group 5 on Linguistic Data and Terminology of the EMT network: the localisation of an existing LLM evaluation dataset into the European languages, to be carried out by master's students across EMT member universities as part of their professional training. This design, embedding a high-impact language technology task within an authentic academic workflow, defines the dual aim of the initiative (see Introduction). The first phase of the project is scheduled for completion by 30 June 2026, in alignment with the spring academic semester.

### 3.2. The MMLU Dataset Translation Project

The MMLU dataset (Hendrycks et al., 2021) is a widely used benchmark for evaluating LLMs through multiple-choice questions spanning diverse domains (e.g., history, law, mathematics, medicine, engineering). As noted earlier, existing multilingual extensions of MMLU have relied predominantly on MT or crowdsourced MTPE, approaches that have been shown to introduce translation artefacts, terminological inconsistencies, and culturally inappropriate renderings that can distort evaluation outcomes (Singh et al., 2024). While such approaches offer scalability, their suitability for high-stakes benchmarking contexts remains an open question.

Extending the MMLU dataset to European languages addresses gaps in linguistic representation and supports more equitable evaluation in the context of the European Commission's and DGT's active role in the evaluation of LLMs in all of the EU, focusing on ensuring linguistic and cultural diversity across Europe pursuant to the overarching principle of multilingualism and the principle of 'no language left behind'. The goal of the European Commission is aligned with the Digital Decade Policy Programme 2030, in actively pursuing "Digital Language Equality" (DLE). The strategy focuses on using LLMs and AI to bridge the widening gap between English and other European languages. In this vein, there has been growing evidence that such multilingual LLMs exhibit strong capabilities for some or many of the under-resourced languages they have been trained in (Rehm, Grützner-Zahn & Barth, 2025:1). The EMT network substantially backs up this EU-wide effort by actively participating in this project with significant impact given that it has members in almost all EU member states representing many under-resourced languages in its programmes.

The dataset comprises around 15,000–16,000 items (over 1,400,000 words in English). Its concise, structured format makes it well suited to collaborative translation workflows at scale.

### 3.3 Project Design and Participation Model

In this initiative, we opted for full human translation, without MT as a basis, carried out by master's students in translation under expert supervision. Human translation remains the 'golden standard' for multilingual benchmarking and cross-lingual evaluations as MT generated data can introduce biased unnatural phrasing (Artetxe, Ruder & Yogatama, 2020:5). This methodological choice reflects a commitment to semantic fidelity, terminological consistency, and cultural appropriateness as preconditions for the valid benchmarking of multilingual systems. In addition, by not relying on neural machine translation (NMT) for this purpose —even when followed by human post-editing— we avoid a recursive feedback loop in which automatically generated content is reused to train or evaluate systems, sometimes described as the 'ouroboros effect' (Shumailov et al., 2024).

The EMT network is composed of 81 master's programmes across 24 countries. All EMT programmes and their students were invited to participate in the localisation of the dataset. Each programme can engage in the project according to the modality that best fits its academic planning and constraints.

Three main forms of participation have been established:

- Integration within a course or module, where the entire class group contributes to the project as part of structured teaching activities introduced and supervised by the instructor.

- Participation through remote cooperation agreements coordinated between the university and the DGT.

- Voluntary participation as an extracurricular activity, allowing students to contribute outside the formal curriculum.

The call for participation has been positively received, with 21 EMT university programmes joining, 230 students, among whom 2 student project managers, from 11 EMT member countries actively translating and revising into 11 target languages currently in the project. The volume of words to be translated into each target language is not fixed, but rather dynamically determined by the number of participating institutions and the specific mode of student involvement adopted in each case. This results in a flexible and scalable distribution of tasks across languages and institutions (see Table 1).

| Language Pair | No. of Students | Expected volume (in words) | Universities |
| --- | --- | --- | --- |
| EN–CZ | 14 | 94.400 | Charles University |
| EN–DE | 3 | 70.000 | University of Vienna; University College Cork |
| EN–EL | 4 | 80.000 | Aristotle University of Thessaloniki |
| EN–ES | 10 | 82.000 | Universitat Autònoma de Barcelona; Universidad de Salamanca; University of Lisbon; University College Cork |
| EN–FR | 27 | 149.300 | Université Bourgogne Europe; Université de Liège; University College Cork; University of Lisbon |
| EN–HU | 11 | 78.200 | Eötvös Loránd University; University of Vienna |
| EN–IR | 1 | 20.000 | University College Cork |
| EN–IT | 100 | 704.000 | University of Bologna; Università degli Studi Internazionali di Roma – UNINT; Katholieke Universiteit Leuven; University of Trieste; Università degli Studi di Udine; Université Bourgogne Europe; University College Cork; University of Lisbon; University of Naples “L’Orientale” |
| EN–NL | 3 | 50.000 | Leiden University |
| EN–PL | 39 | 202.000 | University of Warsaw; University of Krakow (UKEN); University of Gdańsk; University of the National Education Commission, Krakow University |
| EN–PT | 18 | 44.010 | University of Lisbon; Université Bourgone Europe |
| **TOTAL** | **230** | **1.574.910** | |

Table 1: Overview of language pairs, number of participating students, expected word volume, and participating universities

In parallel, the DGT has formalised remote cooperation agreements with those students who commit to translating and in turn revising at least total of 20,000 words, thereby aligning the project with recognised forms of professional training and ensuring adequate institutional support and supervision. These remote cooperation agreements between DGT and the involved EMT programmes —which constitute standard practice for other academic projects involving an agreement between DGT and a university and carried out remotely— contribute to strengthening the link between academic training and professional practice within the EMT framework. In total 11 agreements were signed with participating universities, involving 35 students. This level of participation reflects both the interest generated by the initiative and its capacity to mobilise a substantial number of remote trainees across different educational and linguistic contexts.

### 3.4 Project Management and Coordination

Project management is coordinated within the EMT framework by Work Group 5 on Linguistic Data and Terminology, led by Pilar Sánchez-Gijón (Universitat Autònoma de Barcelona) and Susana Valdez (Leiden University). In close collaboration with the AI Unit of the Directorate-General for Translation (DGT) and EMT officials, the general conditions for project execution were defined, including workflow design, quality standards, and coordination mechanisms across participating institutions.

At the operational level, responsibility for managing the translation process is assigned to two student project managers (PMs), each representing one of the coordinating master’s programmes (Leiden University and Universitat Autònoma de Barcelona). These students act as project managers, creating and assigning tasks, ensuring consistency and quality, facilitating communication by addressing queries and resolving technical issues and serving as the main point of contact between their local teams and the central coordination structure. This model reinforces both project management skills and accountability, while ensuring alignment across languages and institutions.

In addition, the project benefits from the systematic support of the DGT’s Language Units, which act as expert advisors throughout the workflow. They provide authoritative guidance on terminological decisions, domain-specific conventions, and language-specific nuances, ensuring that translations align with interinstitutional standards and remain consistent across languages. To operationalise this alignment, an ad hoc style guide was developed for the project, drawing on the DGT’s interinstitutional style guides and selectively consolidating the conventions most relevant to the MMLU dataset. The guide, prepared by a student from the Universitat Autònoma de Barcelona, serves as a practical reference that harmonises terminology, punctuation, formatting, and QA criteria across teams, thereby enhancing usability and coherence. Beyond its immediate function, this framework exposes students to professional revision practices and establishes a structured interface between academic training and institutional expertise.

The project was formally launched on 16 February 2026 with a kick-off meeting bringing together all participating universities, marking the official start of activities and coordination across institutions.

### 3.6 Workflow and Task Execution

Following the initial kick-off meeting, detailed information has been collected regarding the volume of words that each student commits to translating and revising. Based on this allocation, participants were granted access to a dedicated RWS Trados Enterprise tenant managed by the Universitat Autònoma de Barcelona, through which all project management and translation activities are carried out.

Each student undertakes a dual role, committing to both the translation and revision of an equivalent volume of text. PMs create specific projects for each language pair and assign each file to one student to be translated or revised. These files typically contain a variable volume of text, ranging from just over 100 words to approximately 1,800 words, allowing for manageable yet substantial units of work.

Participants receive task notifications directly through the platform, which are also reinforced via email communication to ensure timely awareness and engagement. In parallel, faculty members who have integrated the project into their courses are provided with PM-level access, enabling them to monitor progress, retrieve assignments, and support their students throughout the process.

To facilitate communication and problem-solving, a dedicated Microsoft Teams environment was set up within the DGT tenant, bringing together all project participants. Within this space, a shared Q&A spreadsheet is maintained, where students systematically record doubts and issues encountered during translation. The PMs oversee this process, providing initial responses and, when necessary, escalating queries to the relevant DGT Language Units. This structured communication channel ensures efficient knowledge sharing, traceability of decisions, and consistency across languages.

## 4 Key Challenges and Operational Constraints

This section outlines the main challenges encountered in the design and implementation of the project, which stem from its scale, multilingual scope, and institutional complexity. These challenges are not only operational but also methodological and pedagogical, as they directly impact the reliability of the dataset as an evaluation resource, the coordination across a distributed academic network, and the learning experience of participating students. The following subsections discuss these challenges from three complementary perspectives: the construction of a European multilingual evaluation dataset, the administrative and coordination framework required to sustain the initiative, and the practical execution of large-scale collaborative translation workflows. Since the project is ongoing, this section reflects an initial reflection on the topic.

### 4.1 The challenge of building a European evaluation dataset

A central methodological issue lies in the US-centric bias present in parts of the MMLU dataset, which becomes apparent in a number of its questions. Direct adaptation of these items to a European context during localisation would risk modifying their difficulty and, in turn, compromising cross-lingual comparability and construct validity. To address this, a selective two-step approach has been adopted. Items with a global scope or without culturally bound references will be translated directly without content adaptation, while those containing local-specific elements undergo a prior revision process to neutralise or adapt such references to a European setting before translation. This strategy aims to preserve both semantic equivalence and item difficulty across languages.

Another methodological issue lies in the fact that some parts of the MMLU dataset feature translated quotes from sources in European languages, yielding the question whether to find the source text or to retranslate the quote from English to the relevant language (French, for example). In this case, the team decided to ask the translators to find the source of the quote to ensure accuracy of the testing material.

Even with these measures in place, the development of a multilingual evaluation dataset at the European scale entails further methodological constraints. Maintaining comparability requires careful control of semantic precision, difficulty calibration, and domain coverage.

### 4.2 Administrative and coordination challenges

The administrative dimension follows a staged, adaptive coordination model. In a first phase, the coordinators compiled a consolidated overview of participating institutions, student contributors, projected translation volumes, and the designated academic representatives authorised to sign on behalf of each university. This mapping supports a calibrated engagement strategy. Formal agreements (when in place) have been subsequently established with universities whose projected contribution approximates the workload of a standard remote internship, thereby ensuring alignment between academic participation and recognised professional training frameworks. Upon completion, each student will be issued an official certificate of collaboration specifying the number of words translated and revised, providing formal recognition of their contribution for both academic and professional purposes.

### 4.3 Execution and workflow challenges

At the operational level, project execution raises challenges related to scale, quality assurance, and the coordination of geographically distributed teams. The workflow follows a structured, student-led translation–revision model in which each participant translates a set of files and reviews an equivalent volume produced by peers. For language combinations with particularly high workload capacities—most notably Polish, French and Italian—a second revision layer has been introduced to strengthen quality control and ensure greater consistency.

A key responsibility of PMs is the fine-grained calibration of task allocation, taking into account individual student availability as well as, where relevant, the constraints imposed by course-based participation. This requires ongoing adjustments to maintain both feasibility and an equitable distribution of work across participants.

Finally, the management of Q&A processes constitutes a critical coordination layer. It demands continuous monitoring and rapid response times to avoid delays in the workflow. A centralised Q&A system is shared across all language projects, enabling the systematic identification of recurring issues and promoting the efficient dissemination of solutions across language pairs. This approach reinforces cross-lingual consistency and decision traceability.

## 5 Final Remarks

This paper has reported on a large-scale, collaborative localisation project aimed at extending the MMLU benchmark to 11 European languages within the EMT–DGT framework. Beyond its technical objectives, the initiative functions as a structured training environment that brings together academic institutions, students, and EU language professionals in a shared workflow. The project demonstrates the potential of coordinated, multilingual efforts to contribute to more inclusive and representative evaluation resources for LLMs, while simultaneously fostering professional competences aligned with the EMT framework (EMT Expert Group 2022).

At the same time, the project is conceived as a learning process in itself. For the purposes of this presentation, feedback will be collected from both participating students and faculty members, as well as from the DGT's translation units involved in the project, in order to better understand its impact and identify areas for improvement. These perspectives will provide valuable insights into the pedagogical value of the initiative, the effectiveness of the workflow design, and the challenges encountered in practice, thereby informing future iterations and potential scaling of the model.

Beyond its immediate scope, the project also highlights its potential for replicability and transferability to other domains and datasets. The creation of evaluation resources that reflect European cultural, social, and linguistic realities is of clear strategic interest, particularly in the context of developing more inclusive and representative language technologies. Equally significant is the collaborative model underpinning this initiative, which leverages the distributed effort of a large number of participants to achieve outcomes that would be difficult to attain individually. This collective approach not only enhances efficiency and coverage, but also creates valuable opportunities for training, knowledge exchange, and experiential learning within a real-world, multilingual setting.

## Acknowledgments

The authors would like to express their sincere gratitude to all participating students, as well as to the professors and programme coordinators from the EMT network, for their commitment and active involvement in this initiative. Their collaboration has been essential to the development and success of the project.